\documentclass[letterpaper, 10 pt, conference]{ieeeconf}  %

\IEEEoverridecommandlockouts                              %

\usepackage{calc}
\usepackage{cite}
\usepackage{mathptmx}
\usepackage[T1]{fontenc}

\usepackage{amsmath}
\usepackage{amssymb}

\usepackage{graphicx}
\usepackage{wrapfig}
\usepackage{multicol}
\usepackage{booktabs}
\usepackage{multirow}
\usepackage[export]{adjustbox}
\usepackage{array}
\usepackage{colortbl}

\usepackage[table,xcdraw]{xcolor}

\let\labelindent\relax
\usepackage{enumitem}

\usepackage[noend]{algorithm2e}
\usepackage{algcompatible}

\makeatletter
\let\NAT@parse\undefined
\makeatother

\usepackage[hidelinks]{hyperref}

\definecolor{lightblue}{rgb}{0.21,0.49,0.74}
\newcommand{\pol}[1]{\pi_{#1}}

\title{\LARGE \bf
ChairNav: Cross-Embodiment Pre-training and Personalization for Long-Horizon Wheelchair Navigation
}

\author{Junhua Huang$^{1,*}$\thanks{$^{*}$Equal contribution.}, Zhizheng Liu$^{1,*}$, Honglin He$^{1}$,
Xiang Zhang$^{1,2}$, and Bolei Zhou$^{1}$\\[4pt]
\small $^{1}$University of California, Los Angeles\quad
$^{2}$Louisiana State University\\[3pt]
\small\href{https://vail.cs.ucla.edu/ChairNav/}{\textcolor{blue}{\nolinkurl{https://vail.cs.ucla.edu/ChairNav/}}}
}

\hypersetup{pdftitle={ChairNav: Cross-Embodiment Pretraining and Personalization for Long-Horizon Wheelchair Navigation},pdfauthor={Junhua Huang, Zhizheng Liu, Honglin He, Xiang Zhang, Bolei Zhou}}

\makeatletter
\IEEEaftertitletext{%
  \begin{minipage}{\textwidth}
    \centering
    \includegraphics[width=\textwidth]{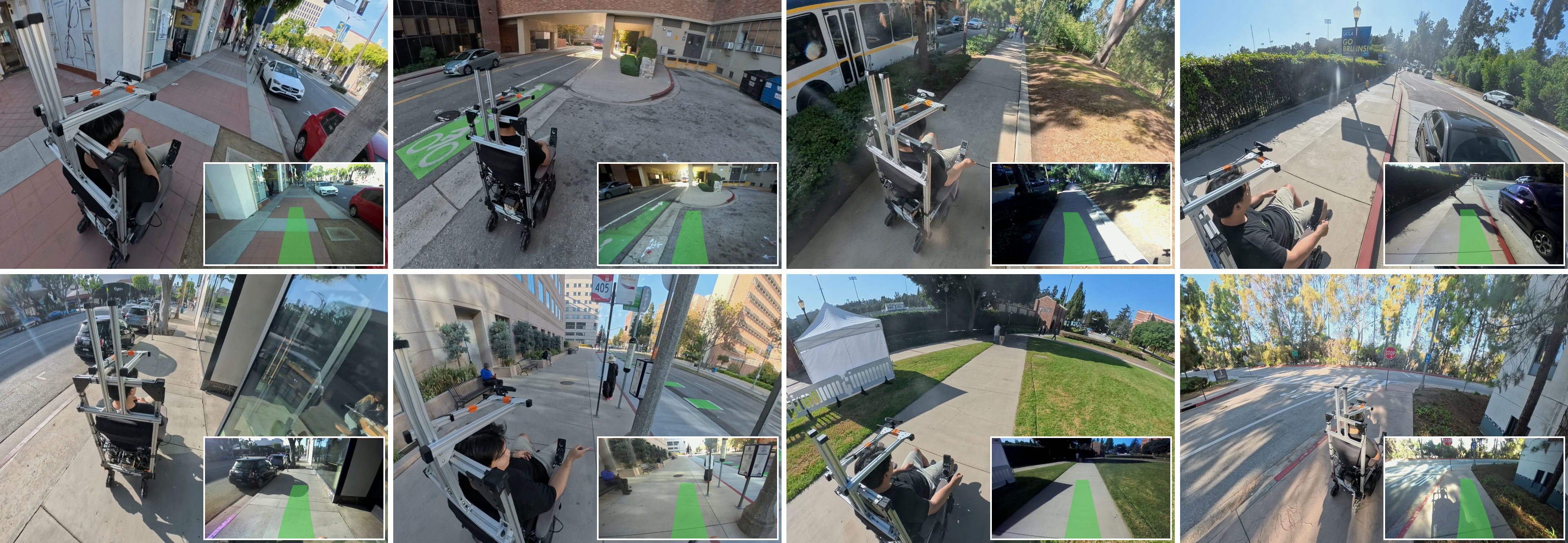}
    \def\@captype{figure}
    \caption{\textbf{Real-world deployment of ChairNav.} Eight autonomous navigation examples on urban sidewalks. Insets show the corresponding first-person views, with model-predicted paths in green.}
    \label{fig:teaser}
  \end{minipage}\vspace{10pt}%
}
\makeatother

\begin{document}

\maketitle
\thispagestyle{empty}
\pagestyle{empty}

\begin{abstract}

Developing autonomous electric wheelchairs capable of long-horizon sidewalk navigation is critical for accessible personal mobility, yet existing approaches either focus on indoor map-based navigation or rely on expensive, complex sensor stacks to achieve full autonomy.
In this work, we propose to tackle outdoor sidewalk navigation by leveraging visual navigation foundation models and large-scale navigation datasets, which offer strong generalization using lightweight hardware.
To this end, we present ChairNav, a novel framework that adapts a general visual navigation model into a personalized electric wheelchair navigation system using a three-stage data pyramid.
To address the scarcity of in-domain data, ChairNav first pre-trains on large-scale cross-embodiment datasets transformed into the wheelchair domain, followed by mid-training on a small number of wheelchair demonstrations to eliminate remaining domain gaps. Finally, a post-training stage refines the deployed model from user interventions to personalize its behavior.
We deploy ChairNav on a commercial electric wheelchair equipped with an RGB-D camera and an onboard NVIDIA Jetson compute module.  Experimental results validate the effectiveness of each stage in our data pyramid: pre-training reduces lateral ADE by $18\%$, and post-training increases the mean distance between user interventions from $29.2\text{ m}$ to $57.4\text{ m}$. Further evaluation on kilometer-scale urban sidewalk routes shows that our system operates autonomously for $93.4\%$ of total driving time, demonstrating its potential for everyday accessible mobility. Demo video and more details are at \url{https://vail.cs.ucla.edu/ChairNav/}. %

\end{abstract}

\section{INTRODUCTION}

Electric wheelchairs (EWs) provide essential mobility assistance to individuals with physical impairments. Safely operating an EW requires constant human attention and maneuvering skills for tasks such as obstacle avoidance and reactive navigation on cluttered sidewalks, posing significant challenges for users with visual, motor, or cognitive impairments. Therefore, developing autonomous EW systems that navigate challenging sidewalk conditions and reduce the burden of manual control is critical to accessible mobility.

Extensive research on autonomous EWs has focused on shared-control systems that mitigate safety risks while providing driving assistance~\cite{udupa2023shared,carlson2012collaborative}. However, these systems still require active, continuous control inputs from the user. 
Although a few studies~\cite{schwesinger2017smart,cui2025research} have tackled  autonomous EW driving, they predominantly rely on traditional modular pipelines equipped with laborious and expensive sensor stacks, which often suffer from integration complexity and error accumulation across modules. 
Conversely, recent advances in end-to-end visual navigation foundation models~\cite{shah2023gnm,shah2023vint,liu2025citywalker,he2026learning} and large-scale visual navigation datasets~\cite{liu2025citywalker, suomela2026data} offer a promising alternative with a lightweight sensor stack and strong generalization ability. Nevertheless, directly deploying these models on EWs or fine-tuning them with existing datasets yields suboptimal performance due to significant gaps in sensor placement, platform dynamics, and physical constraints. 
Meanwhile, collecting EW demonstrations for imitation learning (IL) is also prohibitively expensive, posing challenges to training the EW navigation model.

Furthermore, since EWs are personal mobility devices, their autonomous operation must align with individual users' preferences, such as maintaining a specific distance from the sidewalk curbs, adjusting comfortable driving speeds, or navigating a neighborhood that differs significantly from the data used to train the models. Yet current paradigms rely on fixed models after deployment, failing to continuously adapt to user feedback and improve the user experience during real-world operation.

In this work, we present ChairNav, a novel learning framework that adapts visual navigation foundation models for autonomous EW navigation. To bridge the domain gap of existing models and datasets, we introduce a data transformation technique that aligns cross-embodiment observations and dynamics with the target EW domain via viewpoint reprojection and trajectory re-fitting. This enables us to leverage existing large-scale visual navigation datasets to effectively train visual navigation foundation models. In addition, to accommodate individual driving styles, we develop an efficient preference learning strategy that  handles delays in user takeovers and refines the model through deployment-time user feedback.
Consequently, our framework forms a data pyramid to train the models from three complementary stages:
1) \textit{Pre-training} on 170 hours of transformed cross-embodiment data, 2) \textit{Mid-training} on 3.19 hours of high-quality wheelchair demonstrations to bridge the embodiment gap, and 3) \textit{Post-training}  using 0.48 hours of human preference feedback to align the user's habits and expectations.

Our system is built upon a commercial EW with an RGB-D camera to provide observations and an NVIDIA Jetson computer with a CAN-bus interposer that lets the low-level control system receive control commands from both the model and the user for preference learning (Fig.~\ref{fig:teaser}). Extensive experiments demonstrate that the pre-training stage is crucial for achieving robust navigation: compared to fine-tuning the base model with only the wheelchair demonstrations, it reduces lateral ADE from 0.082~m to 0.067~m. Concurrently, the post-training stage continuously aligns the system with user preference, increasing the average distance between user interventions from 29.2~m to 57.4~m.
In addition, indoor evaluations of autonomous EWs are short-horizon, with reported routes typically under 50~m~\cite{burhanpurkar2017cheap}.
Instead, we evaluate on kilometer-scale drives on real-world urban sidewalks, and the results show that our final system operates  autonomously for 93.4\% of the total driving time in urban public spaces, underscoring its strong potential to improve the user's everyday mobility.

Our main contributions are summarized as follows:
1) ChairNav, a data-pyramid learning framework that adapts visual navigation foundation models for autonomous long-horizon sidewalk navigation.
2) A data transformation method that aligns cross-embodiment data with the target EW domain to address the lack of in-domain data. Compared to fine-tuning the base model on only the wheelchair data, the transformed data reduces lateral ADE by 18\%.
3) An efficient preference learning method for post-training the model during real-world deployment. Post-training on only 0.48~h of user takeovers raises the distance between interventions by 97\%.

\section{RELATED WORK}

\subsection{Autonomous wheelchair navigation}

Prior work on autonomous wheelchairs has predominantly adopted shared-control paradigms that keep the user in the loop~\cite{veterans2005smart,sivakanthan2022mini}, either by filtering continuous joystick commands~\cite{levine1999navchair} or predicting user intent and intervening when necessary~\cite{carlson2012collaborative}. Systems that achieve autonomous and kilometer-scale navigation typically rely on modular pipelines and environment-specific information prepared in advance, such as large-scale landmark maps for over $12$~km of outdoor navigation~\cite{schwesinger2017smart}, prior 3D maps for $8.8$~km of Tokyo sidewalks~\cite{niijima2019real}, GPS/RTK-GNSS waypoints for routes of about 1~km~\cite{fujimoto2010autonomous,nguyen2022enabling}, or routes obtained from commercial map services~\cite{cui2025research}. Other evaluations are largely restricted to short or controlled settings, including doorway traversal~\cite{burhanpurkar2017cheap}, a $100$~m hospital route~\cite{takahashi2024safety}, imitation-learned person following~\cite{salimpour2025follow}, and approximately $100$~m of visual servoing along tactile paving~\cite{hafez2025lightweight}; indeed, a survey of $132$ robotic wheelchairs found only five designed for both indoor and outdoor operation~\cite{sivakanthan2022mini}. In contrast, our approach learns an end-to-end visual wheelchair navigation model that requires neither continuous user control nor a prior map, survey pass, or pre-specified waypoint route, and we evaluate it over $1.61$~km of public urban sidewalks using only a forward-facing camera, with closed-loop results reported in Table~\ref{tab:drives}.

\subsection{Visual navigation foundation models and datasets}

Recent work learns generalizable visual navigation foundation models from large-scale heterogeneous data spanning robots, environments, and embodiments. GNM combines eight datasets from six robot platforms~\cite{shah2023gnm}, ViNT scales this paradigm to hundreds of hours of navigation experience~\cite{shah2023vint}, and NoMaD unifies goal-directed navigation and exploration within a diffusion policy~\cite{sridhar2024nomad}. Navigation data have been collected through teleoperation~\cite{karnan2022socially}, autonomous robot operation~\cite{hirose2023sacson}, and human-carried sensors~\cite{nguyen2023toward}, while recent efforts further scale to $2{,}000$~h of web-sourced walking videos~\cite{liu2025citywalker}, crowdsourced teleoperation relabeled by a model-based expert~\cite{hirose2025learning}, and $4{,}565$~h of teleoperation across $35$ countries~\cite{suomela2026data}. For sidewalk navigation, \textsc{MIMIC}~\cite{he2026learning} and FlowPilot~\cite{he2026imitation} leverage teleoperated robot fleets, while recent models have demonstrated up to $125$~km of map-free navigation~\cite{suomela2026data} and further improvement through reinforcement-learning post-training~\cite{he2026seeing,chen2026socialnav}. 
Instead, our work focuses on adapting these models to autonomous wheelchairs, which present a substantial domain gap and limited in-domain data. Additionally, we aim to align the model with individual user preferences to enable effective personalization.

\section{METHOD}
\label{sec:method}
\begin{figure*}[t]
  \centering
  \includegraphics[width=\textwidth]{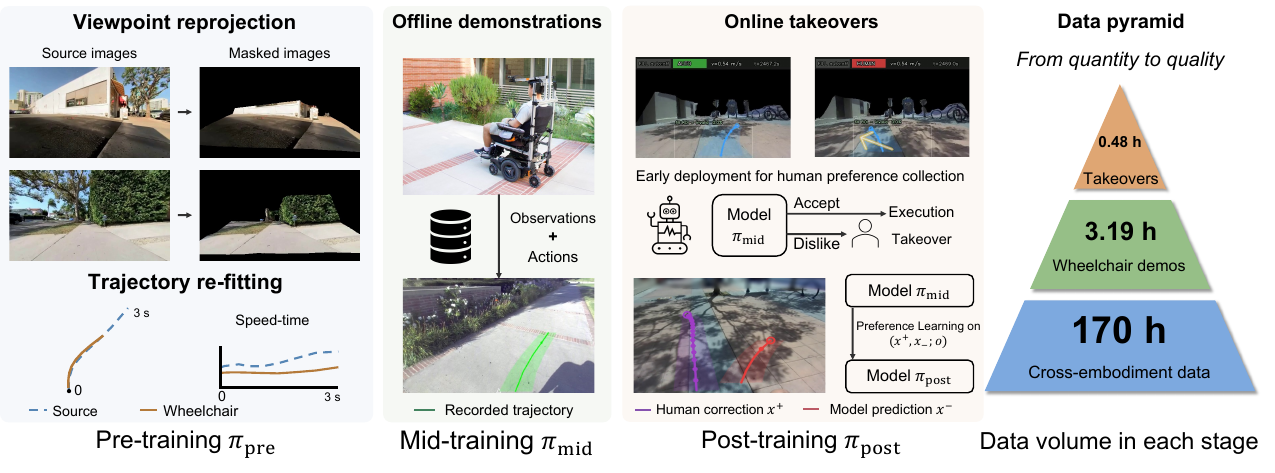}
  \caption{\textbf{Overview of ChairNav.} Cross-embodiment data, wheelchair demonstrations, and user takeovers are utilized in three successive adaptation stages. In the pre-training stage, source data are transformed to align with the wheelchair platform, and black masks computed by viewpoint reprojection restrict the source images to the wheelchair camera's visual coverage. In the post-training stage, users may override the model whenever its behavior does not match their driving preferences. In post-training, $x^{+}$ denotes a corrective target derived from takeover feedback and $x^{-}$ the model prediction for the corresponding pre-takeover observation $o$. The pyramid on the right-hand side represents the amount of training data at each stage; tier sizes are schematic. 
  }
  \label{fig:overview}
\end{figure*}

\subsection{Overview}
Our goal is to develop an autonomous wheelchair navigation system by leveraging massive visual navigation data and navigation foundation models, as hundreds of hours of sidewalk navigation video demonstrations have been collected by food delivery robot companies like Coco Robotics, and various capable sidewalk navigation models~\cite{he2026seeing, he2026learning, he2026imitation} have been trained with strong generalization ability across environments and embodiments.

However, transferring these models and datasets to a wheelchair is challenging because camera viewpoints and vehicle dynamics differ substantially between wheelchairs and delivery robots. These differences affect both the input visual observations and the output trajectories represented in the training data. We therefore propose to transform the visual data to match the wheelchair's camera setup, train the model, and further refine the model through user feedback collected after deployment to accommodate individual driving preferences.

We propose ChairNav, a three-stage learning framework organized as a data pyramid as shown in Fig.~\ref{fig:overview}. In the pre-training stage (Sec.~\ref{sec:pretrain}), we transform large-scale cross-embodiment navigation data through viewpoint reprojection and trajectory re-fitting to account for the wheelchair's visual coverage and motion constraints, and use the transformed data to fine-tune a pre-trained navigation model. In the mid-training stage (Sec.~\ref{sec:midtrain}), we further fine-tune the model on human-driven wheelchair demonstrations to address the remaining domain gaps in the actual sensor and vehicle configuration. Finally, in the post-training stage (Sec.~\ref{sec:posttrain}), we use supervision from user takeovers during test time to align the model with individual driving preferences.

\subsection{Pre-training: viewpoint reprojection and trajectory re-fitting}
\label{sec:pretrain}
\begin{figure}[t]
  \centering
  \includegraphics[width=\columnwidth]{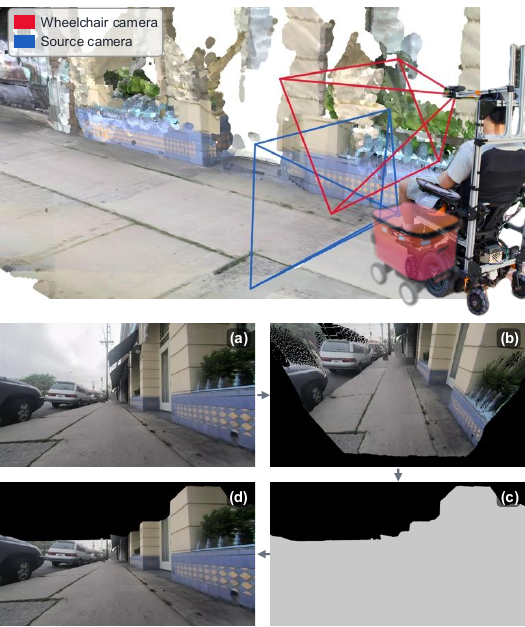}
  \caption{\textbf{Source-data viewpoint reprojection.} Top: a reconstructed scene with source (blue, $0.44$~m) and wheelchair (red, $1.47$~m, $27.7^\circ$ downward) camera frusta. The source data is collected by the pink food delivery robot shown in the image. Bottom: (a) source image; (b) projection into the wheelchair view; (c) visibility mask in source-image coordinates (grey: visible); (d) masked source image used for training.}
  \label{fig:reprojection}
  
\end{figure}

In the pre-training stage, a base visual navigation foundation model $\pi_{\text{base}}$ is fine-tuned on large-scale visual navigation demonstrations collected using a fleet of four-wheeled delivery robots, and we transform both the observations and trajectories to align with the wheelchair platform. To bridge the viewpoint gap between the two setups, we first calibrate the relative transformation $
\mathbf{T}$ from the delivery robot's front camera to the camera we installed on our wheelchair platform. Next, for each image in the training dataset, we compute a visibility mask $\mathbf{M}$ indicating the pixels visible from the wheelchair's camera perspective. Let $\mathbf{K}_{\mathrm{s}}$ and $\mathbf{K}_{\mathrm{w}}$ denote the intrinsic matrices for the source and wheelchair cameras, respectively. A source pixel $\mathbf{u}$ with depth $d(\mathbf{u})$ estimated by a monocular metric-scale depth estimation model~\cite{lin2025depth} can be reprojected to the wheelchair camera frame as:
\begin{equation}\label{eqn:project}
\mathbf{u}_{\mathrm{w}} = \Pi( \mathbf{K}_{\mathrm{w}}, \mathbf{T}\cdot\Pi^{-1}\left(\mathbf{K}_{\mathrm{s}},\mathbf{u}, d(\mathbf{u})\right )),\end{equation} 
where $\Pi$ and $\Pi^{-1}$ represent the perspective projection and unprojection.
The mask $\mathbf{M}(\mathbf{u})$ is thus obtained by setting $\mathbf{M}(\mathbf{u}) = 1$ for pixels whose projected coordinates fall within the image dimensions of the wheelchair camera with positive depths, and $0$ otherwise. 
We then use these masked source images as transformed observations. Compared to reprojecting the source images into the wheelchair's viewpoint, masking leaves the source pixels unwarped, so the training images carry no reprojection artifacts. It also introduces little overhead during model inference, as we can use the depth provided by the RGB-D camera on the wheelchair platform to warp the image to the source viewpoint using $\mathbf{T}^{-1}$ to obtain the masked input. Figure~\ref{fig:reprojection} illustrates the source-data transformation used for training. The source dataset consists of 170 hours of video footage of small-sized food delivery robots being teleoperated along sidewalks~\cite{he2026learning}.

We also adjust the output trajectory to ensure it remains within the wheelchair's kinematic limits. To be specific, we fit an approximate nonlinear kinematic model from calibration runs, capturing its velocity and acceleration bounds as well as turning responses, including inertial and slope-dependent effects. Using the original source trajectory as a reference, we simulate the wheelchair's motion under these identified dynamics, and the resulting rollouts serve as the adapted trajectory targets.

The resulting transformed dataset consists of the masked source observations $\mathbf{o}$ paired with re-fitted trajectories $\mathbf{x}$ and their speed profiles $\mathbf{v}$, and we use it to fine-tune the base model FlowPilot~\cite{he2026imitation}.
Following~\cite{he2026imitation}, we adopt anchored flow matching with the following losses:
\begin{equation}
\mathcal{L}_{\mathrm{BC}}=\mathcal{L}_{\text{FM}}+\mathcal{L}_{\text{CE}}+\mathcal{L}_{\text{speed}},
\label{eq:pre}
\end{equation}
where $\mathcal{L}_{\mathrm{FM}}$ is the flow matching loss, $\mathcal{L}_{\mathrm{CE}}$ is the anchor classification loss over the $K=64$ trajectory anchors, and $\mathcal{L}_{\mathrm{speed}}$ penalizes motion-speed error. Please refer to~\cite{he2026imitation} for the loss and training details. Minimizing Eq.~\eqref{eq:pre} yields the pre-trained model $\pi_\text{pre}$.

\subsection{Mid-training: imitation on wheelchair driving}
\label{sec:midtrain}
While the pre-training data are transformed to align with the wheelchair setup, a domain gap remains in actual sensor and physical constraints. For instance, a feasible path for a delivery robot may collide with walls or obstacles due to the wheelchair's larger dimensions. Furthermore, wheelchair dynamics support motion maneuvers---such as sharp turns or turning in place---that are rarely present in the pre-training dataset.
Therefore, we collect a small dataset comprising 3.19 hours of human wheelchair demonstrations to fine-tune the model during mid-training.
Our mid-training objective combines the imitation learning loss $\mathcal{L}_{\text{BC}}$ with a KL regularizer that anchors the model  $\pi_\theta$ to its pre-trained initialization $\pi_\text{pre}$:
\begin{equation}
\mathcal{L}_{\mathrm{mid}}
=\mathcal{L}_{\mathrm{BC}}
+\lambda_{\mathrm{KL}}
D_{\mathrm{KL}}\big(p_{\pi_{\text{pre}}}(\cdot\mid \mathbf{o})\,\big\|\,p_{\pi_\theta}(\cdot\mid \mathbf{o})\big),
\label{eq:mid_loss}
\end{equation}
where $p_{\pi_{\theta}}(\cdot\mid \mathbf{o})$ is the discrete distribution over the trajectory anchors instead of the intractable trajectory probability density. Using the KL constraint, $\pi_\theta$ is penalized for dropping any anchor that $\pi_\text{pre}$
supports, discouraging concentration on the dominant modes of the wheelchair demonstrations. We denote the resulting mid-trained model $\pi_\theta$ as $\pi_\text{mid}$, which is deployable for real-world testing.

\begin{figure}[t]
  \centering
  \includegraphics[width=\columnwidth]{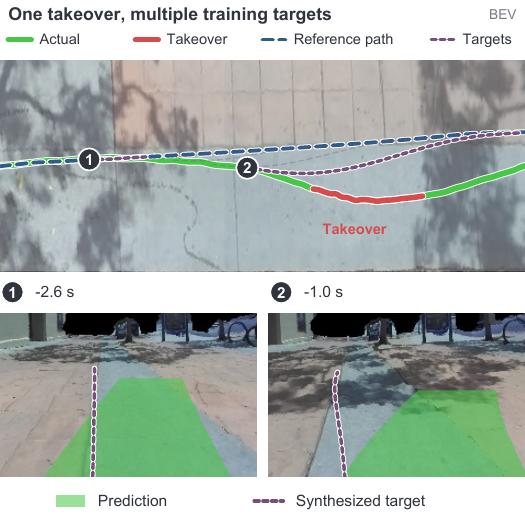}
  
  \caption{\textbf{Preference synthesis from one takeover.} Top: bird's-eye view of the path driven (green), the user's takeover (red), and the reference path (blue dashed) recovered from the takeover. Observations 1 and 2, taken $2.6$~s and $1.0$~s before the takeover, each receive a synthesized target (purple dashed) that leads onto the reference path. Bottom: the camera view at the two observations with the model's prediction (green) and the synthesized target.}
  \vspace{-3mm}
  \label{fig:ppl}
\end{figure}

\subsection{Post-training: learning from user interventions}

\label{sec:posttrain}

To align the model with the user's preferences, we collect user interventions during real-world deployment, where the user can take over at any moment through the joystick and overwrite the model prediction with the corrective action. We record $1{,}117$ takeover events, totaling $0.48$~h of user-controlled driving during takeovers.

However, learning directly from raw interventions presents a significant challenge: preferences are inherently scarce, noisy, and subject to temporal misalignment. Recent interactive preference learning methods~\cite{cai2026predictive} address preference scarcity by projecting interventions forward into future rollouts, assuming the robot and human will maintain consistent actions over a future preference horizon. While effective in simulation, forward preference augmentation breaks down on physical wheelchairs: human takeovers are inherently delayed by system latency, mechanical momentum, and human reaction time.  Instead of predicting future preferences, our key insight is to augment historical states by retroactively synthesizing corrective targets anchored to the post-intervention recovery path.

User dissatisfaction may accumulate over time before prompting a takeover. Inspired by~\cite{cai2026predictive}, we associate each takeover with a preceding preference horizon and apply the user's corrective feedback retrospectively to observations within this horizon, as illustrated in Fig.~\ref{fig:ppl}. For each valid takeover, we automatically generate a reference path using a procedure derived from manually annotated ideal paths. Each eligible observation $\mathbf{o}$ is assigned a corrective target $\mathbf{x}^{+}$ through a quintic spline joining its recorded pose to the reference path, matching position, heading, and curvature at both ends. Among the candidate landing points, we choose the nearest one whose spline respects a curvature bound and let the target follow the reference path thereafter. Each $\mathbf{x}^{+}$ is paired with the model's own prediction $\mathbf{x}^{-}$ at the same observation. This produces multiple training pairs from a single takeover.

Given the augmented preference pairs $(\mathbf{o},\mathbf{x}^{+},\mathbf{x}^{-})$, we initialize $\pi_\theta$ from the mid-trained policy $\pi_{\mathrm{mid}}$ and further refine it using the imitation learning objective together with a reference-constrained preference loss following Direct Preference Optimization (DPO)~\cite{rafailov2023direct}:
\begin{align}
r_\theta(\mathbf{x},\mathbf{o})
&\triangleq
\log \frac{
p_{\pi_\theta}(A(\mathbf{x})\mid\mathbf{o})
}{
p_{\pi_{\mathrm{ref}}}(A(\mathbf{x})\mid\mathbf{o})
},\\
\mathcal{L}_{\mathrm{post}}
&=
\mathcal{L}_{\mathrm{BC}}
-\lambda_{\mathrm{p}}
\log\sigma\!\left(
\beta\left[
r_\theta(\mathbf{x}^+,\mathbf{o})
-r_\theta(\mathbf{x}^-,\mathbf{o})
\right]\right),
\label{eq:post_loss}
\end{align}
where $\sigma(\cdot)$ is the sigmoid function and $\beta$ is a hyperparameter. We set the reference policy to the mid-trained model, i.e., $\pi_{\mathrm{ref}}=\pi_{\mathrm{mid}}$. Similar to the KL regularization used in mid-training, the preference objective is applied only to the anchor classification head, where $A(\cdot)$ maps each trajectory to its nearest anchor. This reference-constrained objective encourages the post-trained policy to favor human-preferred trajectories while limiting its deviation from $\pi_{\mathrm{mid}}$. The resulting model, $\pi_{\mathrm{post}}$, is adapted to wheelchair navigation while preserving the behavior learned during mid-training.

\section{SYSTEM}
 
\label{sec:system}

 \begin{figure}[t]
  \centering
  \includegraphics[width=\columnwidth]{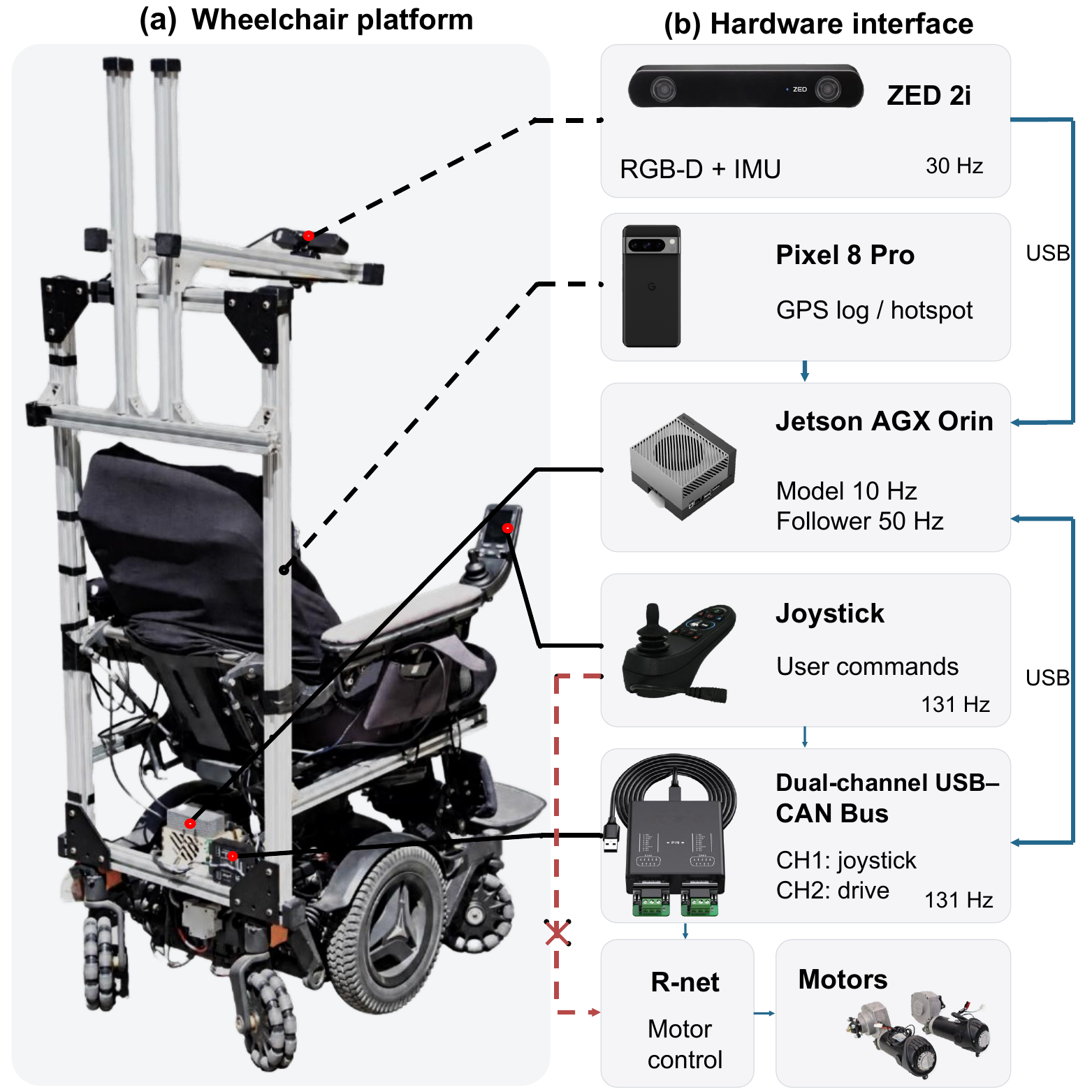}
  \caption{\textbf{Wheelchair platform.} (a) The wheelchair is equipped with the camera mast, the Jetson enclosure and the interposer for autonomous navigation. (b) Hardware interface for communication and controlling: the dual-channel USB--CAN adapter sits between the R-net joystick and the motor controller; OEM modules and programming are unchanged.}
  \label{fig:hardware}
\end{figure}

\subsection{Overview}
\label{sec:chain}
Our system, illustrated in Fig.~\ref{fig:hardware}, is built upon a Permobil M3 power wheelchair. We modify its factory PG Drives R-net control system with an in-line CAN-bus interposer that transparently rewrites joystick position frames to execute model commands in autonomous mode, while also allowing the user to take over via the joystick. A Jetson AGX Orin developer kit
carries all computation, and a dual-channel USB--CAN adapter connects the Jetson to the drive bus.  A Stereolabs ZED~2i RGB-D camera provides the observation and is installed on a mast $1.47$~m above the ground pitched $27.7^\circ$
down.
Odometry comes from the camera's visual--inertial tracking as the chair has no
wheel encoders, and a tethered phone
supplies GPS coordinates used only for logging. No factory electronics are replaced, and the chair's R-net configuration is not reprogrammed. This non-invasive interface avoids replacing or reprogramming the OEM modules and provides a practical path toward adaptation to other wheelchairs with compatible control interfaces.

The model's inference frequency is $10$~Hz. The camera captures both the RGB and depth images, which we use for depth reprojection and visibility masking.
As the model only outputs future trajectories, we use a pure-pursuit controller with a $1.5$~m look-ahead to transform the predicted future waypoints into a virtual joystick command at $50$~Hz.

\label{sec:interface}

\subsection{Compensator}
Due to systematic execution errors like unstable chassis dynamics, trajectory-to-joystick mapping noises, and camera-to-body calibration errors, the physical wheelchair trajectory could deviate from the model prediction, causing user interventions that correct execution errors rather than express driving preferences.  
We therefore compensate for predicted trajectories before execution to improve tracking accuracy and reduce execution-related
interventions in the takeover data.
The compensator combines three constant terms accounting for crab angle, yaw-rate drift, and longitudinal under-execution with an IMU and speed-conditioned residual model for slope-related drift. These estimates are used to adjust the predicted trajectories before conversion to joystick commands. With $S_i$ the cumulative arc length of the predicted trajectory at its $i$-th waypoint, $t_i$ its time stamp and $v$ the current speed, each waypoint $\mathbf{p}_i=(x_i,y_i)$ is replaced by $\mathbf{p}_i-(\Delta x_i,\Delta y_i)$ with
\begin{align}
\Delta x_i &= g_x\, k\, v\, t_i, \qquad \\
\Delta y_i &= g_y\left(\sin\theta\, S_i + \tfrac{1}{2}\,\omega\, S_i^{2}/v\right) + g_r\, r_i(\mathbf{z}),
\label{eq:comp}
\end{align}

where $k$ is the longitudinal under-execution ratio, $\theta$ an effective crab angle, $\omega$ a zero-command yaw-rate drift, and $r_i$ a linear residual model on speed and IMU features $\mathbf{z}$.

With the proposed compensator module, we reduce lateral drift on straight segments over the $3$~s  horizon from about $18$~cm to $2$~cm, and the proportion of takeovers attributed to execution errors also decreases from $34\%$ to $10\%$, thus making takeover data more informative about user preferences.
The fitted gains are close to one, but we deliberately apply only part of the correction ($g_y=0.6$, $g_r=0.5$) and cap the lateral correction at the $3$~s planning horizon at $0.4$~m. These limits are intended to reduce overcompensation, which can push the wheelchair off the sidewalk where recovery can be difficult and dangerous. The correction is applied only while the model is driving at $v\geq0.3$~m/s and the plan reaches at least $1.5$~m ahead; it is low-pass filtered over $0.3$~s and faded in and out over $0.5$~s, so that a change in the plan never becomes a sudden lateral jump.

\section{EXPERIMENTS}

We first describe the implementation details of ChairNav and the evaluation metrics in Sec.~\ref{sec:impl_detail} and Sec.~\ref{sec:metrics}.
We then report open-loop evaluation results to show the effectiveness of pre-training
and mid-training to the general wheelchair navigation performance in Sec.~\ref{sec:cross_adapt} and  post-training for aligning user preferences in Sec.~\ref{sec:learing_inter}.
We finally conduct long-horizon  real-world experiments in Sec.~\ref{sec:real_world_exp}.

\subsection{Model implementation details}
\label{sec:impl_detail}

Mid-training uses $3.19$~h of human wheelchair driving data collected across different environments such as campus, residential areas, and commercial streets, and an additional $0.61$~h of data is held out for validation and open-loop evaluation of the general wheelchair navigation performance. All parameters are fine-tuned for four epochs with $\lambda_{\mathrm{KL}}=10$, and the model after the second epoch is taken as $\pi_{\text{mid}}$.

Post-training data come from five drives of a single user, $4.96$~h in total with $1{,}117$ takeover events. We hold out $121$ takeover events for evaluation and use the remaining takeovers to synthesize corrective targets as described in Sec.~\ref{sec:posttrain}.

We exclude takeovers at intersections and crosswalks, detours, and stops longer than $5$~s. For preference synthesis, the horizon spans $4$~s. Reference paths are generated automatically as described in Sec.~\ref{sec:posttrain} and extended along their final heading when needed. The curvature bound for the joining spline is $0.25$~m$^{-1}$.

All open-loop results are computed at the deployment operating point: deterministic ODE integration with four Euler steps, the top-scoring anchor as the prediction, and the same masked input as at deployment.

\subsection{Evaluation metrics}
\label{sec:metrics}
For open-loop evaluation, we use
average displacement error (ADE) which measures the mean Euclidean
distance between the predicted trajectory waypoints and the ground truth, and final
displacement error (FDE) which measures their distance at the end of the prediction horizon (3~s).

At fixed prediction times, ADE reflects differences in both forward progress and lateral position. Since lateral alignment determines whether the wheelchair stays on the sidewalk and clears obstacles, we additionally report lateral ADE (LADE), defined as the mean absolute difference between the lateral coordinates of predicted and reference waypoints, expressed in the wheelchair's coordinate frame at prediction time.

For Table~\ref{tab:offline_eval}, we group windows by the
user's lateral displacement at $3$~s, with positive values
denoting leftward motion: left turns
($y_{\mathrm{GT}}\geq0.10$~m, $n=641$), straight motion
($|y_{\mathrm{GT}}|<0.10$~m, $n=836$), and right turns
($y_{\mathrm{GT}}\leq-0.10$~m, $n=708$).
Table~\ref{tab:offline_eval} averages errors over windows.
For Table~\ref{tab:intervention}, we first average errors
within each takeover event and then average across events,
giving each event equal weight.

For closed-loop evaluation, a takeover is a continuous episode of user control; consecutive takeovers separated by less than $2$~s of model control are counted as one takeover, and the whole span counts as user control time. Autonomy is the percentage of driving time under model control. Distance per takeover is the total traveled distance divided by the number of takeovers. We also report takeover counts by correction type: leftward, rightward, and braking.

\subsection{Cross-embodiment adaptation}
\label{sec:cross_adapt}

We evaluate cross-embodiment adaptation with pre-training and mid-training on the general wheelchair validation set in Table~\ref{tab:offline_eval}.
The comparison between $\pi_\text{base}$ and $\pi_\text{pre}$
shows that pre-training on transformed source data improves
lateral prediction over the base model without using wheelchair
demonstrations. Subsequent results with $\pi_\text{mid}$ further reduce LADE
across left-turn, straight, and right-turn windows.

To underscore the importance of pre-training, we also directly
fine-tune $\pi_\text{base}$ on wheelchair demonstrations used in mid-training without pre-training. This variant, $\pi_{\text{mid-only}}$, achieves the lowest overall ADE and FDE, indicating closer matching of the
forward progress over the prediction horizon. However, its turning
trajectories are less accurate laterally: left- and right-turn
LADE are $0.100/0.113$~m, compared with $0.083/0.097$~m for
$\pi_{\text{mid}}$. Thus, lower overall displacement error does not
translate into more accurate turning trajectories.

Pre-training improves lateral accuracy in all three motion
categories, reducing pooled LADE from $0.082$ to $0.067$~m
relative to $\pi_{\text{mid-only}}$, an improvement of approximately $18\%$.

\subsection{Learning user preference}
\label{sec:learing_inter}

\begin{figure*}[!t]
  \centering
  \includegraphics[width=\textwidth]{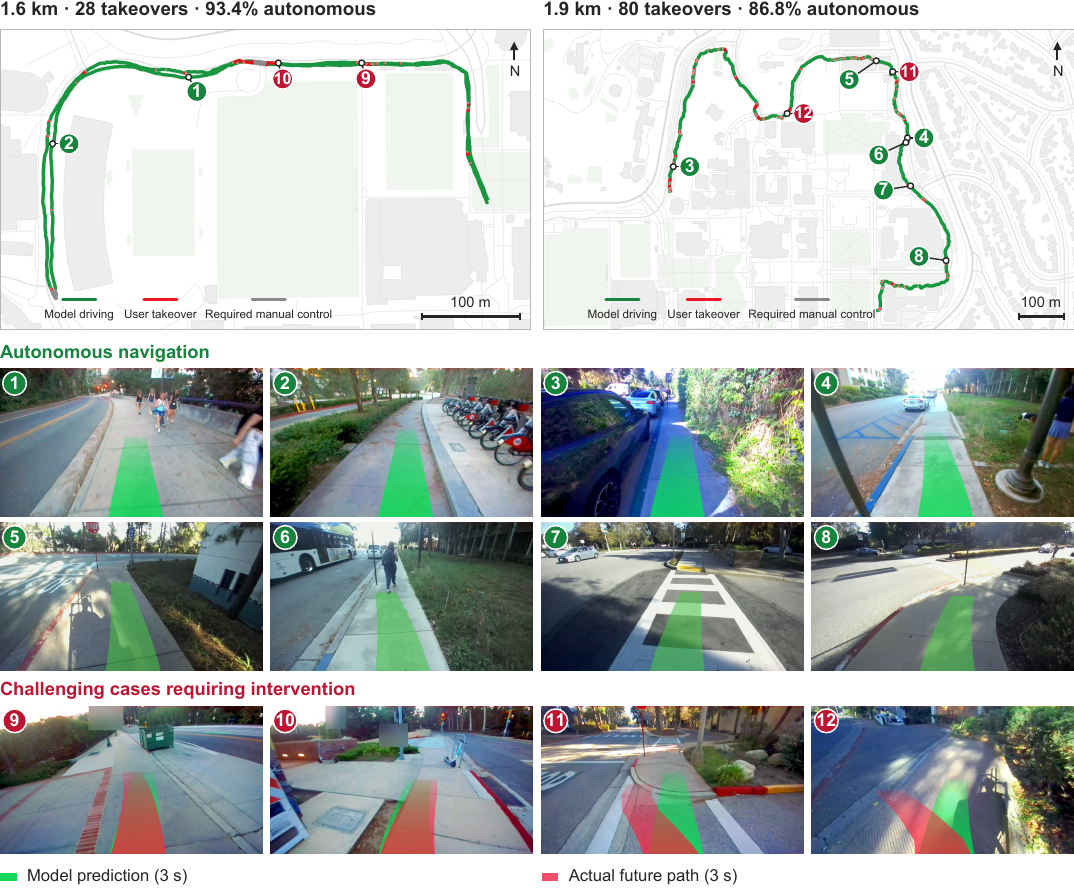}
    \caption{Long-horizon navigation with $\pol{\text{post}}$ on $1.6$ and $1.9$~km routes, with $93.4\%$ and $86.8\%$ of driving time under model control. GPS tracks show model driving (green), user takeovers (red), and required manual control (grey); dots mark takeover onsets. Camera views show eight autonomous navigation examples (1--8), followed by four challenging cases requiring intervention (9--12). Green bands show the model's $3$~s plans, swept by the wheelchair's $0.72$~m width; red bands show the path actually driven over the next $3$~s. The second track ends when onboard compensation was switched off. Basemap \copyright~OpenStreetMap contributors. Faces are blurred.}
  \label{fig:long_horizon}
\end{figure*}

We evaluate the contribution of post-training by comparing $\pi_\text{mid}$ and $\pi_\text{post}$ on the user validation set with  an ablation
of post-training without the  synthesized
preference labels (w/o pref. synthesis). Both $\pi_\text{post}$ variants are initialized from
$\pi_{\text{mid}}$ and trained with the same human takeover data.
As shown in Table~\ref{tab:intervention}, including synthesized
preference labels reduces ADE from $0.208$ to $0.161$~m,
FDE from $0.514$ to $0.484$~m, and LADE from $0.072$ to
$0.069$~m compared with the ablated variant. The resulting
$\pol{\text{post}}$ also achieves lower errors than $\pol{\text{mid}}$ across
all three metrics. These results show that post-training with synthesized
preference labels improves agreement with the user's preference. This gain does not come at the expense of general driving accuracy: on the general validation set (Table~\ref{tab:offline_eval}), $\pi_{\text{post}}$ matches $\pi_{\text{mid}}$ in lateral error (LADE $0.066$ vs.\ $0.067$~m) with a small increase in ADE ($0.237$ vs.\ $0.213$~m).

\subsection{Real-world closed-loop deployment}
\label{sec:real_world_exp}

We evaluate $\pol{\text{mid}}$ and $\pol{\text{post}}$ on the same $1.61$~km urban
sidewalk route, with three trials per policy. All six trials were
conducted within a single day, alternating between the two
policies to reduce the effects of changing battery condition,
illumination, and pedestrian activity.  
Before every trial, all
hardware was fully shut down and restarted, followed by a
five-minute warm-up to reduce variation in system operation and
inference speed. One trial per policy was recorded in full; Table~\ref{tab:drives} reports these two recorded trials.

Figure~\ref{fig:long_horizon} visualizes the spatial distribution of autonomous control and user interventions during the recorded $\pol{\text{post}}$ trial and during a $1.9$~km $\pol{\text{post}}$ drive on a different route (Scenario B in Table~\ref{tab:all_drives}).
On the maps, green denotes autonomous control and red denotes user control; on the $1.6$~km track, grey marks sections requiring manual control: a crossing where the user operates the pedestrian signal and a designated U-turn. Takeovers in these sections remain included in the reported statistics. Both maps show autonomous operation over most of each route, with user interventions distributed along the sidewalk. The eight autonomous examples cover pedestrians, docked bikes, parked cars, roadside poles, a crosswalk and a curved sidewalk. The four takeovers occurred where $\pol{\text{post}}$ planned toward a sidewalk edge, a planter or a curb.

Table~\ref{tab:drives} summarizes the closed-loop results.
Autonomy increases from $91.8\%$ under $\pol{\text{mid}}$ to $93.4\%$ under $\pol{\text{post}}$, while distance per takeover increases from
$29.2$ to $57.4$~m. The number of takeovers decreases from $55$ to $28$. This reduction is concentrated in leftward
corrections, which decrease from $38$ to $15$;
rightward corrections stay at $12$. Thus,
$\pol{\text{post}}$ requires fewer user interventions,
with a more balanced distribution of lateral correction
directions.

\begin{figure}[t]
  \centering
  \includegraphics[width=\columnwidth]{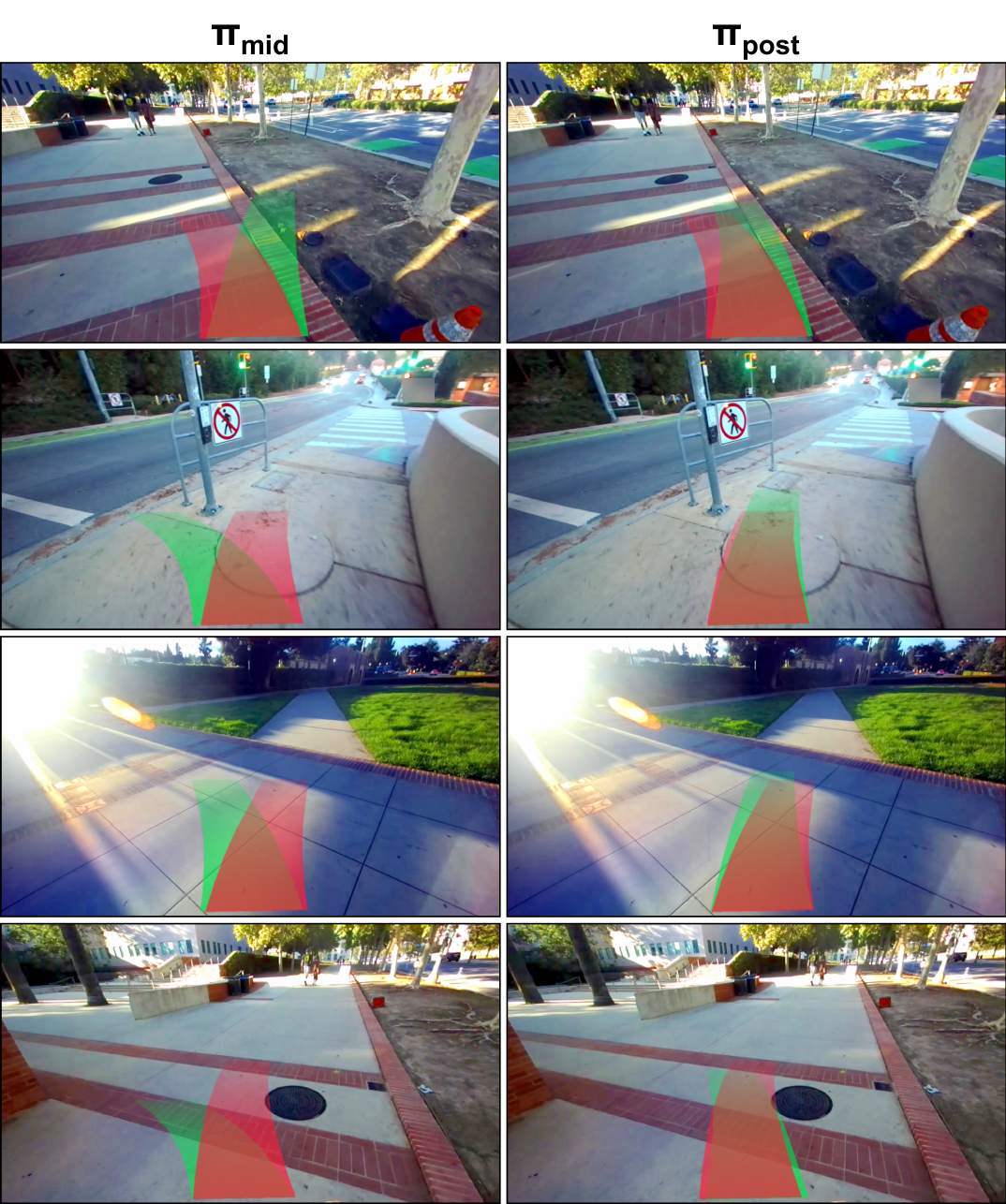}
    \caption{Predicted trajectories of $\pi_\text{mid}$ (left) and $\pi_\text{post}$ (right) on the same recorded inputs at four user takeovers: a tree pit beside a brick walkway, a signal pole at the end of a sidewalk, a sidewalk in low sun, and a brick strip next to a drain cover. Green bands show the top-1 plans swept by the wheelchair's $0.72$~m width; red shows the path the wheelchair actually followed over the next $3$~s. Rows 1, 2 and 4 were recorded under $\pi_\text{post}$ and row 3 under $\pi_\text{mid}$; the other model is evaluated offline on the same recorded input.}
  \label{fig:mid_post}
\end{figure}

Figure~\ref{fig:mid_post} shows four takeovers at which the two models disagree by $0.6$--$1.7$~m at the $3$~s horizon. $\pi_\text{mid}$ plans into the tree pit (row 1), veers left at the signal pole (row 2) and onto the brick strip (row 4), while $\pi_\text{post}$ stays on the walkway. In all four cases the end of $\pi_\text{post}$'s plan lies within $0.45$~m of the path then driven, against $1.0$--$1.5$~m for $\pi_\text{mid}$.

\begin{table}[t]
\caption{Open-loop evaluation on the general wheelchair validation set. We evaluate LADE separately on the left-turn, straight and right-turn segments.  }
\label{tab:offline_eval}
\centering
\scriptsize
\setlength{\tabcolsep}{2pt}
\renewcommand{\arraystretch}{1.15}
\begin{tabular*}{\columnwidth}{@{\extracolsep{\fill}}l ccc cc cccc@{}}
\toprule
& & & & & & \multicolumn{4}{c}{LADE~$\downarrow$} \\
\cmidrule(lr){7-10}
Model & Pre-train & Mid-train & Post-train & ADE~$\downarrow$ & FDE~$\downarrow$ & Left & Straight & Right & All \\
\midrule
$\pi_\text{base}$ &  & & & 0.599 & 1.678 & 0.183 & 0.086 & 0.178 & 0.144 \\
$\pi_\text{pre}$  & \checkmark & & & 0.386 & 0.936 & 0.100 & 0.045 & 0.117 & 0.084 \\
$\pi_\text{mid-only}$ &   & \checkmark & & \textbf{0.152} & \textbf{0.470} & 0.100 & 0.042 & 0.113 & 0.082 \\
$\pi_\text{mid}$ & \checkmark & \checkmark & & 0.213 & 0.546 & 0.083 & \textbf{0.030} & \textbf{0.097} & 0.067 \\
$\pi_\text{post}$  & \checkmark & \checkmark & \checkmark & 0.237 & 0.577 & \textbf{0.078} & \textbf{0.030} & 0.098 & \textbf{0.066} \\
\bottomrule
\end{tabular*}
\end{table}

\begin{table}[t]
\caption{Open-loop evaluation of takeover-based post-training on takeover validation data. Errors are in meters; bold indicates the lowest reported value per column.}
\label{tab:intervention}
\centering
\scriptsize
\setlength{\tabcolsep}{3.5pt}
\renewcommand{\arraystretch}{1.2}
\begin{tabular}{@{}l ccc@{}}
\toprule
  & ADE~$\downarrow$ & FDE~$\downarrow$ & LADE~$\downarrow$ \\
\midrule
\addlinespace[1pt]
$\pi_\text{mid}$& 0.173 & 0.526 & 0.074 \\
\addlinespace[3pt]
$\pi_\text{post}$ (w/o pref. synthesis) & 0.208 & 0.514 & 0.072 \\
$\pi_\text{post}$ & \textbf{0.161} & \textbf{0.484} & \textbf{0.069} \\
\bottomrule
\end{tabular}

\end{table}

\begin{table}[t]
\caption{Closed-loop performance of the recorded trial of each policy on the same $1.61$~km urban sidewalk route.}
\label{tab:drives}
\centering
\scriptsize
\setlength{\tabcolsep}{2.5pt}
\renewcommand{\arraystretch}{1.2}
\begin{tabular*}{\columnwidth}{@{\extracolsep{\fill}}lcccc@{}}
\toprule
Model & \shortstack{Autonomy\\(\%)~$\uparrow$}
& \shortstack{Auto./total\\time (min)~$\uparrow$}
& \shortstack{Takeovers\\(L/R/B)~$\downarrow$}
& \shortstack{Distance per\\takeover (m)~$\uparrow$} \\
\midrule
$\pi_\text{mid}$ & 91.8 & 24.5~/~26.7 & 55 (38~/~12~/~5) & 29.2 \\
$\pi_\text{post}$ & \textbf{93.4} & 25.0~/~26.8 & \textbf{28} (15~/~12~/~1) & \textbf{57.4} \\
\bottomrule
\end{tabular*}
\vspace{2pt}

\vspace{2pt}
\parbox{\columnwidth}{\scriptsize
L/R/B: left/right/brake takeovers.
Values are from the one recorded trial per policy (the first two Scenario A rows in Table~\ref{tab:all_drives}).}
\end{table}

\section{CONCLUSION}
In this work, we present ChairNav, a novel learning framework that adapts general visual navigation foundation models into personalized models for autonomous electric wheelchairs.
Our key idea is to structure the training pipeline into a three-stage data pyramid to address the lack of   in-domain wheelchair data and also  align with individual users'
preferences.
We first transform large-scale delivery robot navigation data to match the wheelchair's visual perspective and motion constraints via viewpoint reprojection and trajectory re-fitting. 
We then mid-train the model on a small set of in-domain wheelchair demonstrations to bridge the remaining embodiment gaps, before post-training on human takeover feedback using latency-compensated preference learning. 
Our system is built upon a commercial electric wheelchair with an additional RGB-D camera and onboard compute. 
Evaluation results highlight the effectiveness of each stage in our data pyramid and ChairNav's strong potential in advancing accessible personal transportation.

\noindent\textbf{Limitations.}
First, our cross-embodiment trajectory re-fitting relies on an approximate wheelchair dynamics model to simulate target rollouts, which does not explicitly verify whether the resulting paths remain physically traversable within the geometry of the original source scenes. Second, real-world human takeovers naturally conflate true driving preference signals with corrective interventions triggered by residual perception noise, camera frame miscalibrations, or execution tracking errors. Because our current post-training objective treats all takeover events uniformly, it does not explicitly disentangle user style preferences from system fault corrections. Finally, our closed-loop real-world evaluations were conducted with a single human user, leaving the model's capacity to personalize across diverse users and environmental conditions unverified.

\bibliographystyle{IEEEtran}
\bibliography{IEEEfull, root}
\begin{figure*}[!t]
  \centering
  \includegraphics[width=\textwidth]{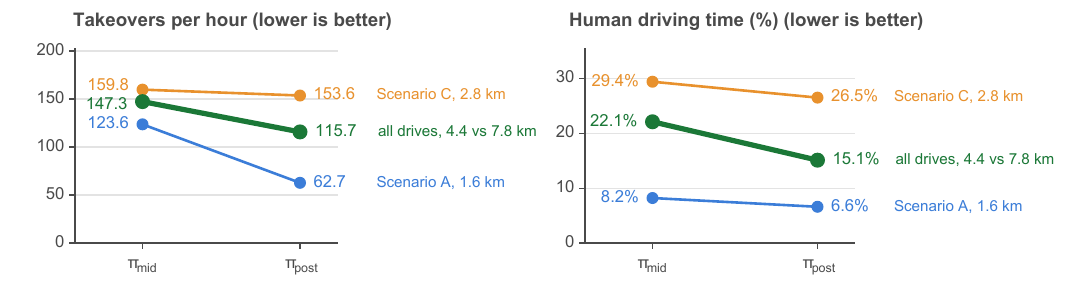}
    \caption{Takeovers per hour (left) and share of driving time under user control (right) of $\pol{\text{mid}}$ and $\pol{\text{post}}$ on the two same-route pairs (Scenario A, $1.6$~km; Scenario C, $2.8$~km) and over all drives ($4.4$~km of $\pol{\text{mid}}$, $7.8$~km of $\pol{\text{post}}$).}
  \label{fig:slope}
\end{figure*}
\newpage
\section*{Appendix}

\subsection*{A. Counting takeovers}
The wheelchair hands control to the user whenever the joystick leaves its dead zone and returns control to the model as soon as the joystick springs back to center. Because the joystick is spring-loaded, a single sustained correction is often delivered as a burst of short deflections separated by fractions of a second of model control, and a raw count of hand-overs overstates the number of distinct interventions. We therefore count consecutive takeovers separated by less than $2$~s of model control as one takeover event, and the whole merged span as user control time.

\subsection*{B. Closed-loop drive records}
Table~\ref{tab:all_drives} lists all closed-loop drives of $\pol{\text{mid}}$ and $\pol{\text{post}}$ with the metrics of Table~\ref{tab:drives}; the last two rows pool all four $\pol{\text{post}}$ drives (Scenarios A--C) and both $\pol{\text{mid}}$ drives (Scenarios A and C). Figure~\ref{fig:slope} plots takeovers per hour and the user-control share for the two same-route pairs and for all drives.

\subsection*{C. Compensator gains}
For safety, the compensator of Sec.~\ref{sec:system} does not apply the full fitted correction. The lateral terms of Eq.~\eqref{eq:comp} are applied with $g_y=0.6$ and $g_r=0.5$. The longitudinal under-execution ratio $k$ was estimated from the calibration logs, but the longitudinal term is not applied ($g_x=0$), because lengthening the commanded path raises the wheelchair's speed, and an unexpected speed-up is harder for the user to recover from than a lateral offset.

\vfill\newpage
\begin{table}[!t]
\caption{Closed-loop drive records.}
\label{tab:all_drives}
\centering
\scriptsize
\setlength{\tabcolsep}{2pt}
\renewcommand{\arraystretch}{1.2}
\begin{tabular*}{\columnwidth}{@{\extracolsep{\fill}}lcccccc@{}}
\toprule
Scenario & Model & \shortstack{Dist.\\(km)} & \shortstack{Autonomy\\(\%)~$\uparrow$}
& \shortstack{Auto./total\\time (min)~$\uparrow$}
& \shortstack{Takeovers\\per km~$\downarrow$}
& \shortstack{Dist. per\\takeover (m)~$\uparrow$} \\
\midrule
A & $\pi_\text{mid}$ & 1.6 & 91.8 & 24.5~/~26.7 & 34.2 & 29.2 \\
A & $\pi_\text{post}$ & 1.6 & 93.4 & 25.0~/~26.8 & 17.4 & 57.4 \\
A & $\pi_\text{post}$ & 1.6 & 95.1 & 25.4~/~26.7 & 18.3 & 54.5 \\
B$^{a}$ & $\pi_\text{post}$ & 1.9 & 86.8 & 29.4~/~33.9 & 42.7 & 23.4 \\
C & $\pi_\text{mid}$ & 2.8 & 70.6 & 35.8~/~50.7 & 47.9 & 20.9 \\
C & $\pi_\text{post}$ & 2.8 & 73.5 & 36.5~/~49.6 & 46.1 & 21.7 \\
\midrule
all & $\pi_\text{mid}$ & 4.4 & 77.9 & 60.3~/~77.4 & 43.0 & 23.3 \\
all & $\pi_\text{post}$ & 7.8 & 84.9 & 116.2~/~136.9 & 33.8 & 29.6 \\
\bottomrule
\end{tabular*}

\vspace{2pt}
\parbox{\columnwidth}{\scriptsize
Scenario A is the route of Table~\ref{tab:drives}. $^{a}$~Up to the point where the onboard compensation was switched off.}
\end{table}
\null

\end{document}